\pdfoutput=1

\documentclass[11pt]{article}

\usepackage{EMNLP2023}

\usepackage{times}
\usepackage{latexsym}

\usepackage[T1]{fontenc}

\usepackage[utf8]{inputenc}

\usepackage{microtype}

\usepackage{inconsolata}
\usepackage{marvosym}
\usepackage{graphicx}
\usepackage{xcolor}
\usepackage{multirow}
\usepackage{tabularx}
\usepackage{float}
\usepackage[ruled,vlined]{algorithm2e}
\usepackage{wrapfig}
\usepackage{lipsum} 
\usepackage{algpseudocode}
\usepackage{subcaption}
\usepackage{amsthm}
\usepackage{enumitem}
\usepackage{amsmath} 
\usepackage{booktabs}
\usepackage{amsfonts}       
\usepackage{nicefrac} 
\usepackage{hyperref}

\usepackage{soul}
\usepackage{dsfont}
\usepackage{xspace}
\newcommand{\ours}{\texttt{BIPEFT}\xspace}

\usepackage{colortbl}
\title{\ours: Budget-Guided Iterative Search for Parameter Efficient Fine-Tuning of Large Pretrained Language Models}

\author{
Aofei Chang$^1$,
Jiaqi Wang$^1$,
Han Liu$^2$\\
\textbf{Parminder Bhatia}$^3$,
\textbf{Cao Xiao}$^3$,
\textbf{Ting Wang}$^4$,
\textbf{Fenglong Ma}$^1$\thanks{Corresponding author.}\\
$^1$Pennsylvania State University,
$^2$Dalian University of Technology\\
$^3$GE Healthcare,
$^4$Stony Brook University\\
$^1$\{aofei, jqwang, fenglong\}@psu.edu,
$^2$hanliu@dlut.edu.cn\\ $^3$\{parminder.bhatia, cao.xiao\}@gehealthcare.com, $^4$twang@cs.stonybrook.edu
}

\begin{document}
\maketitle
\begin{abstract}
Parameter Efficient Fine-Tuning (PEFT) offers an efficient solution for fine-tuning large pretrained language models for downstream tasks. However, most PEFT strategies are manually designed, often resulting in suboptimal performance. Recent automatic PEFT approaches aim to address this issue but face challenges such as search space entanglement, inefficiency, and lack of integration between parameter budgets and search processes. To overcome these issues, we introduce a novel \textbf{B}udget-guided \textbf{I}terative search strategy for automatic \textbf{PEFT} (\ours), which significantly enhances search efficiency. \ours employs a new iterative search strategy to disentangle the binary module and rank dimension search spaces. Additionally, we design early selection strategies based on parameter budgets, accelerating the learning process by gradually removing unimportant modules and fixing rank dimensions. Extensive experiments on public benchmarks demonstrate the superior performance of \ours in achieving efficient and effective PEFT for downstream tasks with a low parameter budget.\footnote{Source code is available at \url{https://github.com/Aofei-Chang/BIPEFT}}

\end{abstract}

\section{Introduction}
Large pre-trained models (PTMs)~\cite{Devlin2019BERTPO,radford2019language} based on Transformer architectures~\cite{vaswani2017attention} have achieved significant success across a variety of downstream tasks through fine-tuning, including applications of healthcare~\cite{ijcai2024p914, luo-etal-2024-corelation}. However, the computational and storage demands of PTMs limit the feasibility of full fine-tuning. To address this, parameter-efficient fine-tuning (PEFT) methods have garnered considerable attention. 
Existing PEFT approaches have demonstrated superior performance on downstream tasks~\cite{xu2023parameter}. However, most methods, such as adapters~\cite{houlsby2019parameter,lin2020exploring,ruckle2021adapterdrop}, BitFit~\cite{zaken2021bitfit}, and LoRA~\cite{hu2022lora,zhang2023adaptive,zi2023delta,zhang2023increlora,dettmers2023qlora} require \textit{manual design} of fine-tuning strategies. Configuring PEFT for different layers of Transformers can result in varying performance outcomes.

\begin{figure}[t]
  \centering
  \includegraphics[width=0.95\linewidth]{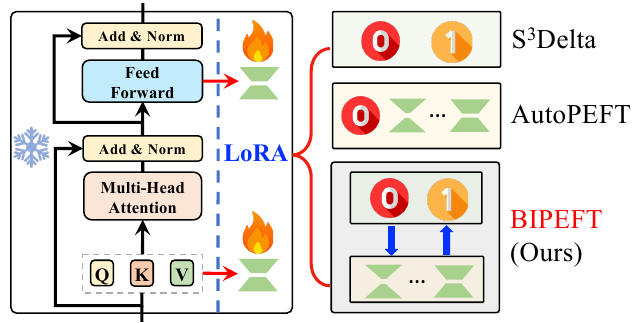}
  \vspace{-0.1in}
  \caption{Search space comparison among different automatic PEFT approaches when searching on low-rank adaption (LoRA)~\cite{hu2022lora}. S$^3$Delta~\cite{hu2022sparse} uses a $\{0,1\}$ binary space to determine whether the module is kept. AutoPEFT~\cite{zhou-etal-2024-autopeft} uses a multi-dimensional space to search module existence and dimension ranks simultaneously, where 0 means the module will be removed. The proposed \ours uses a novel iterative search strategy to disentangle the binary module search and the rank dimension search.}
  \label{fig:search_space_comparison}
\end{figure}

To mitigate this issue, \textbf{automatic PEFT} approaches~\cite{hu2022sparse,zhou-etal-2024-autopeft} have been proposed to automatically search for the optimal PEFT configuration. S$^3$Delta~\cite{hu2022sparse} is the first differential neural architecture search (NAS)-based PEFT approach, which automatically searches for the optimal modules to include in the configuration. AutoPEFT~\cite{zhou-etal-2024-autopeft} employs Bayesian optimization to conduct the PEFT search across a large space. Although these two automatic approaches are effective, they still suffer from the following issues:

\textbf{(1) Search Space Entanglement.} As shown in Figure~\ref{fig:search_space_comparison}, the search space of S$^3$Delta~\cite{hu2022sparse} is limited, focusing only on binary PEFT module searches. In contrast, AutoPEFT~\cite{zhou-etal-2024-autopeft} combines binary module selection with multiple rank dimension spaces by representing binary module selection with 0 and denoting non-zero values for ranks. Such a mixed search ignores the fact that these two spaces are interdependent and entangled. During the search, a module with a rank of 0 means that the search on this module is not necessary yet. In contrast, a non-zero rank indicates that the module will be kept, and 0 will not be selected further. Consequently, jointly searching within such entangled spaces introduces new optimization challenges, requiring an automatic balance between dimensional choices and binary decisions, especially when the search space is large.

\textbf{(2) Better Search Efficiency.} Using a non-differential optimization strategy makes AutoPEFT~\cite{zhou-etal-2024-autopeft} less efficient compared to S$^3$Delta~\cite{hu2022sparse}. However, even with differential NAS, the efficiency of S$^3$Delta is still unsatisfactory due to overlooking the unique characteristics of the PEFT task. Unlike traditional automated machine learning tasks that typically learn model parameters and architectures from scratch, automatic PEFT only learns a small set of parameters, with most being fixed. The weight distributions for some modules or dimension ranks may quickly stabilize after a few training steps. Thus, keeping them involved in the PEFT learning until the final training step is unnecessary and lowers search efficiency.

\textbf{(3) Isolation between Parameter Budgets and Search Efficiency.} In practice, a smaller yet effective model is more useful for downstream tasks. S$^3$Delta~\cite{hu2022sparse} uses a parameter budget to control the number of trainable model parameters. However, this budget does not accelerate PEFT optimization. An ideal solution would be to use the parameter budget as a factor to guide the efficient PEFT configuration, potentially yielding a more effective downstream model.



\begin{figure*}[t]
    \centering
    \includegraphics[width=1\textwidth]{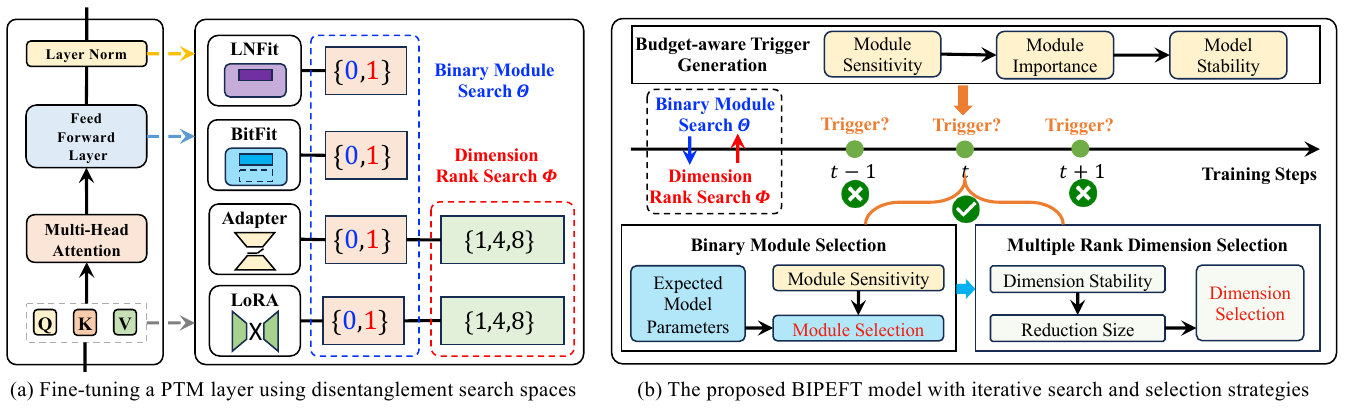}
   \vspace{-0.3in}
    \caption{Overview of the proposed \ours, which conducts an iterative search on disentanglement search spaces with novel module and rank dimension selection strategies to accelerate search efficiency.}
    \label{fig:overview}
    \vspace{-3mm}
\end{figure*}

To solve all the aforementioned issues simultaneously, in this paper, we propose a parameter \textbf{B}udget-guided \textbf{I}terative search strategy \ours for boosting the search efficiency of automatic \textbf{PEFT}, as shown in Figure~\ref{fig:overview}. \ours works as follows: At each training step $t$, \ours uses the designed iterative search strategy to search optimal architecture weights for the binary PEFT module search space and rank dimension search space alternatively, which can handle the first issue. After a few training steps, \ours will trigger the selection modules, where the trigger is generated based on the parameter budget $\mathcal{B}$ and the current module state, as detailed in $\S$Sec.~\ref{method:early-stop}. 
\ours contains a novel parameter budget-guided module selection strategy in $\S$Sec.~\ref{sec:binary_stop} to gradually remove the unimportant modules and a new parameter history-based dimension selection strategy in $\S$Sec.~\ref{sec:dim_stop} to adaptively fix rank dimensions during the search. These two new selection strategies can perfectly address the second and third issues. After selecting modules and fixing rank dimensions, \ours will repeat again until the number of triggers achieves the maximum number $Z$.

To sum up, this work has the following contributions:
\begin{itemize}
    \item We recognize the importance of disentangling the binary module and rank dimension search spaces for automatic PEFT optimization.
    \item We introduce a novel automatic PEFT model \ours, which is an iterative differential NAS-based approach to effectively search for downstream task models with parameter budget constraints.
    \item We design early selection strategies for different space searches, which significantly accelerates the optimal PEFT model learning.
    \item We conduct extensive experiments on two public benchmarks and compare \ours against state-of-the-art baselines. Experimental results demonstrate the efficacy and efficiency of \ours for automatic PEFT.
\end{itemize}

\section{Related Work}




\textbf{Parameter Efficient Fine-Tuning (PEFT).}
Generally, PEFT is designed based on a Transformer architecture and only optimizes a small portion of parameters and leaves the vast majority of parameters frozen for efficient adaptation to downstream tasks. The PEFT methods can be broadly classified into four categories~\cite{han2024parameter}: (1) \textit{Additive PEFT} such as Adapter~\cite{houlsby2019parameter} and Prefix-Tuning~\cite{li-liang-2021-prefix} inserts new trainable modules or parameters in the model. 
(2) \textit{Selective PEFT} aims to optimize model performance by selectively fine-tuning a subset of the model’s parameters by masking~\cite{guo-etal-2021-parameter,fu2023effectiveness,liao-etal-2023-parameter,sung2021training} or manual design~\cite{zaken2021bitfit}.
(3) \textit{Reparameterized PEFT} like LoRA~\cite{hu2022lora},  constructs a reparameterization of the original model parameters for training, then equivalently transforms it back at the inference stage~\cite{zhang2023adaptive, luo2023towards}.
(4) \textit{Hybrid PEFT} focuses on combining the benefits of diverse PEFT modules from the last three categories~\cite{Mao2021UniPELTAU,chen2023parameterefficient}. 
A series of automated hybrid PEFT methods are developed to automatically search for an effective hybrid PEFT structure, and our work falls into this category.

\noindent\textbf{Automated Configuration Search for PEFT.}
AutoPEFT~\cite{zhou-etal-2024-autopeft} using Bayesian optimization and S$^3$Delta~\cite{hu2022sparse} utilizing differential neural architecture search (NAS) techniques to search for optimal architectures for natural language processing (NLP) tasks. In contrast, PrunePEFT \cite{lawton-etal-2023-neural} adopts a straightforward pruning approach to identify essential PEFT parameters. In the vision domain, NOAH~\cite{zhang2022neural} applies an evolutionary NAS strategy to tailor PEFT configurations specifically for Vision Transformer~\cite{dosovitskiy2021an}. Despite these advancements, significant challenges remain in optimizing search space design and improving search efficiency.

\noindent\textbf{Early Stopping in Neural Architecture Search.}
Existing early stopping strategies in neural architecture search address the overfitting issues on selecting many skip connections in convolutional neural networks (CNNs) when using methods like DARTS~\cite{liu2018darts}. OLES~\cite{jiang2023operationlevel} introduces an operation-level early stopping method to mitigate this issue. DARTS+~\cite{liang2019darts+} implements a manual threshold for stopping the search. SGAS-es utilizes an indicator for stabilizing search results on CNNs. Notably, early stopping techniques have not yet been adapted for the PEFT search, which is characterized by faster convergence compared to CNNs.


\section{Methodology}
\vspace{-2mm}
\subsection{Overview}

Our approach aims to automatically search for the optimal PEFT structure from $N$ modules of LLMs through a novel budget-aware search strategy named \ours. This strategy iteratively searches from two distinct spaces: a binary position search space and a dimension search space, with module and dimension selection mechanisms.
As shown in Figure~\ref{fig:overview}, at each training step $t$, \ours will first optimize the architecture weights $\boldsymbol{\Theta}_t \in \mathbb{R}^{N \times 2}$ by fixing the weights of $\boldsymbol{\Phi}_{t-1} \in \mathbb{R}^{N \times K}$ for the dimension search space, where $K$ is the number of dimension candidates. Subsequently, \ours updates $\boldsymbol{\Phi}_{t}$ using the optimized $\boldsymbol{\Theta}_t$. 

After each training step, \ours checks whether the budget-aware selection mechanisms are triggered according to module sensitivity scores and the targeted budget $\mathcal{B}$ detailed in $\S$Sec.~\ref{method:early-stop}. 
If triggered, \ours will first estimate the number of reduced parameters $R_z$ at this trigger stage and then discard unimportant modules according to the estimated $R_z$ with a designed module selection strategy in $\S$Sec.~\ref{sec:binary_stop}. \ours also selects an appropriate dimension size for certain modules in $\S$Sec.~\ref{sec:dim_stop}. These selection operations lead to further updates $\boldsymbol{\Theta}_t$ and $\boldsymbol{\Phi}_{t}$. \ours stops when the maximum trigger count $Z$ is reached, ensuring that the model size approximates the targeted budget $\mathcal{B}$. The search process of \ours is outlined in Algorithm~\ref{algorithm_01}, with the details of the designed strategy discussed in the following subsections.

\begin{algorithm}[t]
\small
\caption{Algorithm of \ours}
\label{algorithm_01}
\KwIn{$N$ PEFT modules with trainable weights $\Delta\mathbf{W}$, architecture parameters $\boldsymbol{\Theta}_0$ and $\boldsymbol{\Phi}_0$, stability trigger threshold $\tau$, parameter budget $\mathcal{B}$, total steps $T$, maximum trigger count $Z$, binary selection indicator $\mathbf{b}_z$, dimension selection indicator $\mathbf{d}_z$}
\KwOut{Searched PEFT architecture $A$}

\For{$t = 1, \ldots, T$}{
    Update $\Delta\mathbf{W}$ by gradient descent\;
    Optimize $\boldsymbol{\Theta}_{t-1}$ and $\boldsymbol{\Phi}_{t-1}$ iteratively according to the objective\;
    Accumulate the sensitivity score $\bar{s}^t$ of each PEFT module using Eq.~\eqref{eq:sensitivity}\;
    Calculate module importance indicator $\mathbf{I}_t$\;
    Evaluate model stability $\beta_t$ using Eq.~\eqref{eq:trigger}\;
    
    \If{$(\beta_t \geq \tau)$}{
        Estimate expected parameters $E_t$ using Eq.~\eqref{eq:expected_parameters}\;
        Calculate expected parameter reduction $R_z$ using Eq.~\eqref{eq:binary_reduction}\;
        Rank $N$ modules by sensitivity score $\bar{s}^t$\;
        Perform early module selection until the reduction $R_z$ is achieved\;
        Update binary selection indicator $\mathbf{b}_z$\;
        Evaluate dimension stability $\lambda_z^n$ using Eq.~\eqref{eq:dsi}\;
        Estimate the number of modules $Y_z$ for dimension selection using Eq.~\eqref{eq:dim_reduction}\;
        Fix dimensions for $Y_z$ modules with the lowest stability scores $\lambda_z^n$\;
        Update dimension selection indicator $\mathbf{d}_z$\;
        Freeze part of the $\boldsymbol{\Theta}_{t-1}$ and $\boldsymbol{\Phi}_{t-1}$ according to selection indicators $\mathbf{b}_z$, $\mathbf{d}_z$\;
        $Z \leftarrow Z-1$\;
    }
    
    \If{$Z \leq 0$}{
        Exit\;
    }
}
\Return{$A$}
\end{algorithm}

\subsection{Budget-aware Trigger Generation}\label{method:early-stop}
A naive solution of automatically searching for the optimal PEFT structure is to gradually reduce the number of parameters during model fine-tuning. However, if the fine-tuned model is unstable, we cannot decide which modules will be pruned. Thus, evaluating model stability is important. To achieve this goal, we design a new budget-aware model stability strategy according to module-level sensitive scores and the targeted budget $\mathcal{B}$.

\subsubsection{Module Sensitivity Estimation}
In our setting, we apply the differential neural architecture search (NAS) for optimal PEFT configuration. If a module $\mathcal{M}_n$ plays an important role in the PEFT model, it should be more stable and contribute greatly to the loss function. In the NAS-based model training, we have both training and validation data, denoted as $\mathcal{D}_{tra}$ and $\mathcal{D}_{val}$. The module sensitivity can be evaluated on these two kinds of data as follows:
\begin{gather}
    s_n^t = f_n^t(\mathcal{D}_{tra}) + \alpha_n^t f_n^t(\mathcal{D}_{val}),\\
    f_n^t(\mathcal{D}) = \frac{1}{|\mathcal{M}_n|}\sum_{w \in \mathcal{M}_n}\Bigl|w\mathbf{G}_n^t(w, \mathcal{D})\Bigr|,\\
    \alpha_n^t = \text{cos}(\mathbf{G}_n^t(\mathcal{D}_{tra}), \mathbf{G}_n^t(\mathcal{D}_{val})).
\end{gather}
Following~\cite{molchanov2019importance}, we use $f_n^t$ to calculate the parameter-level average magnitude of the gradient-weight product. $|\mathcal{M}_n|$ is the number of fine-tuned parameters of the $n$-th module. $\mathbf{G}_n^t(w)$ denotes the gradient of weight $w$. $\alpha_n^t$ denotes the gradient cosine similarity of the module $\mathcal{M}_n$ on both training and validation data. 

The sensitivity score $s_n^t$ of each module can be measured after each training step $t$. To mitigate the impact of stochastic data batch sampling, we propose to smooth the sensitivity score using an exponential moving average following ~\cite{zhang2022platon} as follows:
\begin{equation}\label{eq:sensitivity}
    \bar{s}_n^t = \gamma \bar{s}_n^{t-1} + (1-\gamma) {s}_n^t,
\end{equation}
where $\gamma$ is a predefined hyperparameter.

\subsubsection{Trigger Generation}
Determining the optimal timing for starting parameter reduction is pivotal. To address this challenge, we propose an adaptive trigger generation approach to automatically estimate the optimal timing according to module sensitivity scores learned by Eq.~\eqref{eq:sensitivity} along with the targeted budget $\mathcal{B}$.

\noindent\textbf{Module Importance Indicator.}
Ideally, the finally selected modules should be greatly yet continuously contributed to the model fine-tuning. Moreover, the total parameters of the finally selected modules should be close to the targeted budget $\mathcal{B}$ in our setting. 

Based on these motivations, we design a module importance indicator $\mathbf{I}_t \in \{0, 1\}^N$ for each training step $t$ to record the estimated importance of modules. 
Specifically, we initialize $\mathbf{I}_t = \mathbf{0}$ and rank the module sensitivity scores in descending order. We accumulate the parameters from top-ranked modules one by one. If the current sum is smaller than the targeted budget $\mathcal{B}$, we set the indicator value as 1 for that module. Otherwise, all the remaining indicator values are 0. 

\noindent\textbf{Trigger.} Intuitively, the model performs stably if the important modules change slightly within a time window $H$. We use an average cosine similarity between two consecutive module importance indicators within $H$ steps to evaluate the model stability as follows:
\begin{equation}\label{eq:trigger}
    \beta_t = \frac{1}{H} \sum_{j = t-H}^t \text{cos}(\mathbf{I}_j, \mathbf{I}_{j-1}),
\end{equation}
where $\beta_t \geq \tau$ means that the model is stable now, and the trigger can be generated, where $\tau$ is a hyperparameter. After triggering the parameter reduction, \ours then uses two strategies to reduce the number of training parameters, i.e., binary module selection and multiple dimension selection.

\subsection{Binary Module Selection}\label{sec:binary_stop}
The majority of parameters will be reduced in the binary module selection stage. In the design of \ours, we aim to gradually obtain the optimal PEFT configuration after triggering the reduction $Z$ times. Thus, estimating the number of parameter reductions at the trigger step $z$ is essential. Let $E_t$ denote the expected number of parameters at the $t$-th training step when the trigger counter is $z$. We can estimate the expected number of reductions is 
\begin{equation}\label{eq:binary_reduction}
    R_z = \frac{E_t - \mathcal{B}}{Z-z}.
\end{equation}
Here, the challenge is how can we estimate $E_t$.

\subsubsection{Expected Parameters Estimation}
We use the current status of the PEFT model at the $t$-th training step (i.e., the $z$-th trigger counter) and the selection status at the $(z-1)$-th trigger stage to estimate the expected parameters $E_t$. The PEFT model contains both architecture weights $\boldsymbol{\Theta}_t$ and $\boldsymbol{\Phi}_t$. 
The selection status at $z-1$ contains a binary module selection indicator denoted as $\mathbf{b}_{z-1} \in \{0,1\}^N$ and a module dimension selection indicator vector $\mathbf{d}_{z-1} \in \{0,1\}^N$. $\mathbf{b}_{z-1}^n = 1$ means that the module $\mathcal{M}_n$ is kept in the PEFT training. Otherwise, the module has been removed. $\mathbf{d}_{z-1}^n =1$ indicates that $\mathcal{M}_n$ has a selected or fixed dimension value with index $k_n^*$, but $\mathbf{d}_{z-1}^n = 0$ means the dimension is undetermined.
Based on these notations, we can estimate $E_t$ as follows:
\begin{gather}\label{eq:expected_parameters}
    E_t = \sum_{n=1}^N p_n \sum_{k=1}^K{C}_n^k,
\end{gather}
where $p_n = \mathbf{b}_{z-1}^n * \text{softmax}(\boldsymbol{\Theta}_t^n)[1]$ is the probability of the kept module, and ${C}_n^k$ is the estimated number of parameters of each module, which is defined as follows:
\begin{gather}
    {C}_n^k = \begin{cases}
        \mathbf{q}_n[k_n^*], & \text{if } \mathbf{d}_{z-1}^n =1,\\
        \mathbf{q}_n \cdot \text{softmax}(\boldsymbol{\Phi}_t^n)^\top, & \text{if } \mathbf{d}_{z-1}^n = 0,
    \end{cases}
\end{gather}
where $\mathbf{q}_n \in \mathbb{R}^K$ is the parameter vector for all candidate dimensions, which can be precalculated. When the dimension is fixed, we use the corresponding parameter values directly. Otherwise, we use a weighted sum over the estimated probabilities.

\subsubsection{Adaptive Module Selection}
Using Eq.~\eqref{eq:expected_parameters}, we can obtain the expected reductions $R_z$. To prune unimportant modules, we still use the module-level sensitivity scores calculated by Eq.~\eqref{eq:sensitivity} at step $t$. We rank all sensitivity scores of the currently kept modules and remove the modules with the lowest scores if their total parameter size is smaller than $R_z$. Correspondingly, we update $\mathbf{b}_{z-1}$ to obtain $\mathbf{b}_{z}$ as the new module indicator.

\subsection{Multiple Rank Dimension Selection}\label{sec:dim_stop}
We can also determine the dimension size for each module $\mathcal{M}_n$ if there is a clearly stable pattern on learned architecture weights $\boldsymbol{\Phi}_t^n$. Since evaluating the sensitivity score for each dimension as each module is hard, we propose a new strategy to measure the weight distributions between two trigger counters, $z-1$ and $z$. Note that there are several training steps between the trigger gap. 
Assume that at the $j$-th training step, \ours triggers the reduction $z-1$, and at $t$ ($t>j$), the $z$-th trigger happens. We use the historical architecture weights $[\boldsymbol{\Phi}_j^n, \cdots, \boldsymbol{\Phi}_t^n]$ to evaluate the stability of each dimension.

\subsubsection{Dimension Stability Estimation}
Intuitively, a stable dimension needs to satisfy two conditions. On the one hand, the intra-dimension weight at each training step may not change much, i.e., the standard deviation of dimension-level weights $[\sigma_{z,1}^n, \cdots, \sigma_{z,K}^n]$ should be as small as possible. On the other hand, the cross-dimension weights, i.e., the weight distributions, should also be as similar as possible. Specifically, we can use the Kullback-Leibler (KL) divergence to evaluate the similarity of two distributions. 
Based on these motivations, we design a dimension stability indicator as follows:
\begin{equation}\label{eq:dsi}
    \lambda_z^n = \frac{1}{K}\sum_{k=1}^K \sigma_{z,k}^n \text{KL}(\boldsymbol{\Phi}_j^n, \boldsymbol{\Phi}_t^n).
\end{equation}


\subsubsection{Adaptive Dimension Selection}
Similar to the module selection, we need to determine which modules' dimensions should be selected automatically according to the calculated dimension stability scores $[\lambda_z^1, \cdots, \lambda_z^N]$ using Eq.~\eqref{eq:dsi}. However, the challenge here is estimating the expected number of modules for dimension reduction. 

\noindent\textbf{Expected Reduction Module Size Estimation.}
Intuitively, we can fix more modules' dimensions if the potential dimension selections at $z-1$ and $z$ are similar, which motivates us to use the similarity score to potential dimension vectors, i.e., $\mathbf{v}_{z-1}$ and $\mathbf{v}_z$. The potential dimension of the $n$-th module that is not fixed can be obtained through its architecture weights, i.e., $\mathbf{v}_z^n = \textsf{Dim}[\text{argmax}_{k}(\boldsymbol{\Phi}_t^n)]$, where $\textsf{Dim}$ is all the possible dimension vector. Note that if the module is removed, then $\textbf{d}_{z-1}^n=1, \mathbf{v}_z^n = 0$, and $\mathbf{v}_z^n = \textsf{Dim}[k_n^*]$ for the fixed module. Finally, we can estimate the number of modules for dimension reduction as follows:
\begin{equation}\label{eq:dim_reduction}
    Y_z = \Big\lfloor \frac{\text{sum}(\mathbf{1}-\textbf{d}_{z-1}) * \text{cos}(\mathbf{v}_{z-1}, \mathbf{v}_z)}{Z-z} \Big\rfloor.
\end{equation}

\noindent\textbf{Adaptive Dimension Selection.} After obtaining the expected number of reduction modules, we then use the dimension stability scores $[\lambda_z^1, \cdots, \lambda_z^N]$ to select dimension-unfixed modules among the top $Y_z$ lowest scores, whose dimensions are then fixed using the corresponding values in $\mathbf{v}_z^n$. Finally, we can update $\mathbf{d}_z$ based on $\mathbf{d}_{z-1}$ and the newly dimension-fixed modules.

\begin{table*}[!t]
\centering
\caption{Results on the GLUE and SuperGLUE benchmarks. ``Ratio'' specifies the ratio of trainable parameters compared to T5. Average metrics (AVG) and parameter ratios (Ratio) are averaged over all the values. $*$ denotes the results that are directly copied from \textbf{S$^3$Delta}~\cite{hu2022sparse}. 
}
\label{tab:results}
\vspace{-0.1in}
\resizebox{1\textwidth}{!}{
\begin{tabular}{l|l|ccccccc|c} 
\toprule 
\multicolumn{10}{c}{\cellcolor{gray!10}\textbf{GLUE}} \\\hline
     \textbf{Ratio}
        & \textbf{Method} & \textbf{CoLA} & \textbf{SST2} & \textbf{MRPC} & \textbf{QQP} & \textbf{STSB} & \textbf{MNLI} &  \textbf{QNLI} & \textbf{AVG} \\
        \midrule
        
    10000\%\% & Fine-tune$^*$  & 62.25 ± 3.96 & 95.87 ± 0.42 & 91.86 ± 1.19 & 89.50 ± 0.22 & 91.86 ± 0.46 & 89.61 ± 0.30 & 94.22 ± 0.35 & 87.88 \\ 
    \midrule

    \multicolumn{10}{c}{\textbf{Manual PEFT Methods}} \\\hline
    
    65.33\%\% & Adapter$^*$  &  59.03 ± 3.06 & 95.90 ± 0.29 & 93.02 ± 0.28 & 88.39 ± 0.06 & 91.77 ± 0.25 & 89.53 ± 0.07 & 94.17 ± 0.19 & 87.40  \\ 
    
    21.32\%\% & LoRA($r$=8)$^*$ &  58.43 ± 4.16 & 95.79 ± 0.27 & 92.21 ± 0.88 & 88.35 ± 0.25 & 91.78 ± 0.31 & 89.38 ± 0.32 & 94.14 ± 0.12 & 87.15 \\ 
    
    8.13\%\% & BitFit$^*$  &  56.98 ± 3.89 & 96.24 ± 0.33 & 92.16 ± 0.68 & 88.12 ± 0.07 & 91.59 ± 0.08 & 89.10 ± 0.09 & 94.07 ± 0.21 & 86.90 \\ 
    
    1.70\%\% & LNFit$^*$  & 56.15 ± 4.06 & 95.81 ± 0.20 & 91.71 ± 0.39 & 88.17 ± 0.10 & 91.37 ± 0.24 & 89.11 ± 0.09 & 93.99 ± 0.20 & 86.62 \\ 
    \midrule
    
    \multicolumn{10}{c}{\textbf{Automated PEFT Methods}} \\\hline


     18.90\%\% & AutoPEFT &  59.59 ± 1.24 & 95.87 ± 0.23 & 91.06 ± 0.52 &88.21 ± 0.04 & 91.42 ± 0.08 & 89.16 ± 0.10 & 93.90 ± 0.15 & 87.03  \\ 
     \cellcolor{green!20}1.39\%\% &\cellcolor{green!20}\ours &  \cellcolor{green!20}59.04 ± 8.20 & \cellcolor{green!20}95.70 ± 0.08 & \cellcolor{green!20}92.10 ± 0.61 & \cellcolor{green!20}88.28 ± 0.09 & \cellcolor{green!20}91.83 ± 0.40 & \cellcolor{green!20}89.14 ± 0.01 & \cellcolor{green!20}94.06 ± 0.01 &  \cellcolor{green!20}\textbf{87.16} \\

    \midrule
    
     1.39\%\% & $\text{S}^3$Delta-M$^*$& 59.34 ± 4.75 &95.84 ± 0.14 &92.13 ± 2.09& 88.04 ± 0.23 & 91.58 ± 0.25 &89.14 ± 0.13 &94.12 ± 0.12& 87.17 \\
     1.39\%\% & PrunePEFT & 60.39 ± 5.90 & 95.87 ± 0.09 &92.83 ± 1.41 & 88.12 ± 0.11 & 91.62 ± 0.61 & 89.19 ± 0.15 & 93.95 ± 0.08& 87.42 \\
     \cellcolor{red!20}1.39\%\% & \cellcolor{red!20}\ours &  \cellcolor{red!20}62.97 ± 3.72 & \cellcolor{red!20}96.27 ± 0.24 & \cellcolor{red!20}93.22 ± 0.62 & \cellcolor{red!20}88.28 ± 0.02 & \cellcolor{red!20}92.23 ± 0.28 & \cellcolor{red!20}89.28 ± 0.02 & \cellcolor{red!20}94.25 ± 0.06 & \cellcolor{red!20}\textbf{88.07} \\

 \end{tabular}
}

\resizebox{1\textwidth}{!}{
\begin{tabular}{l|l|cccccc|c} 
\toprule 
\multicolumn{9}{c}{\cellcolor{gray!10}\textbf{SuperGLUE}} \\\hline
\textbf{Ratio} & \textbf{Method}
    & \textbf{BoolQ }& \textbf{CB} & \textbf{MultiRC} & \textbf{ReCORD} & \textbf{RTE} &  \textbf{WIC} & \textbf{AVG}  \\
    \midrule

    10000\%\% & Fine-tune$^*$  &  86.67 ± 0.21  & 96.43 ± 2.92 &  76.65 ± 1.01 & 85.03 ± 0.67 & 88.49 ± 2.12 & 73.12 ± 1.71 & 84.40 \\
     \midrule
     
    \multicolumn{9}{c}{\textbf{Manual PEFT Methods}} \\\hline
     65.33\%\% & Adapter$^*$   & 85.98 ± 0.68 & 94.64 ± 6.19 & 77.60 ± 0.84 & 85.96 ± 0.37 & 89.21 ± 2.94 & 71.63 ± 0.90 & 84.17\\
    
     21.32\%\% & LoRA($r$=8)$^*$ &  85.06 ± 0.70 & 91.96 ± 3.42 & 76.94 ± 1.16 & 85.84 ± 0.21 & 87.05 ± 0.59 & 72.10 ± 1.31 & 83.16   \\

     21.32\%\% & BitFit$^*$ &   85.02 ± 0.48 & 89.29 ± 2.92 & 75.79 ± 1.15 & 85.85 ± 0.32 & 86.15 ± 1.48 & 72.34 ± 1.61 & 82.41 \\

     1.70\%\% & LNFit$^*$  &  84.07 ± 0.50 & 82.14 ± 2.92 & 75.52 ± 1.16 & 86.14 ± 0.11 & 86.69 ± 1.81 & 69.28 ± 1.49 & 80.64 \\

    \midrule
     \multicolumn{9}{c}{\textbf{Automated PEFT Methods}} \\\hline


     17.69\%\% & AutoPEFT &  83.39 ± 0.13 & 93.75 ± 4.49 & 76.49 ± 0.41 & 85.59 ± 0.13 & 85.99 ± 0.42 & 72.17 ± 1.19 & 82.89 \\ 
     
     \cellcolor{green!20}1.39\%\% & \cellcolor{green!20}\ours &  \cellcolor{green!20}84.55 ± 0.13 &  \cellcolor{green!20}92.86 ± 3.58 & \cellcolor{green!20}75.90 ± 0.12 & \cellcolor{green!20}85.94 ± 0.05 & \cellcolor{green!20}88.85 ± 1.53 & \cellcolor{green!20}70.38 ± 0.66 & \cellcolor{green!20}\textbf{83.08} \\
    \midrule
    
     1.39\%\% & $\text{S}^3$Delta-M$^*$ & 84.92 ± 0.68 & 92.86 ± 2.92 & 76.38 ± 0.92 & 86.10 ± 0.11 & 86.69 ± 1.90 & 71.63 ± 1.07 & 83.10 \\
     1.39\%\% & PrunePEFT &  85.16 ± 0.47 & 91.66 ± 4.12 & 76.66 ± 0.60 & 85.46 ± 0.81 & 87.29 ± 0.42 & 70.37 ± 0.23& 82.77 \\
     
     \cellcolor{red!20}1.39\%\% &\cellcolor{red!20}\ours &\cellcolor{red!20}85.44 ± 1.12 & \cellcolor{red!20}94.65 ± 2.52  & \cellcolor{red!20}75.84 ± 0.04  & \cellcolor{red!20}85.70 ± 0.07 & \cellcolor{red!20}88.49 ± 2.04 & \cellcolor{red!20}72.25 ± 0.22 & \cellcolor{red!20}\textbf{83.73} \\ 
     

\bottomrule 
\end{tabular}
}
\vspace{-0.1in}

\end{table*}

\subsection{Iterative Optimization}
\label{method:optim}
The proposed \ours is a differential NAS-based PEFT model, which can be optimized as DARTS~\cite{liu2018darts}. The optimization objective is defined as follows:
\begin{gather*}\small
    \min_{\boldsymbol{\Theta}^*, \boldsymbol{\Phi}^*} \mathcal{L}_{\text{val}}\left(\mathcal{D}_{val};\boldsymbol{\Theta}, \boldsymbol{\Phi}, \mathbf{W}_0+ \Delta\mathbf{W}^*\right), \\\small
    \textit{s.t.\;} \Delta\mathbf{W}^* = \arg\min_{\Delta\mathbf{W}} \mathcal{L}_{\text{tra}}\left(\mathcal{D}_{tra}; \boldsymbol{\Theta}^*, \boldsymbol{\Phi}^*, \mathbf{W}_0+ \Delta\mathbf{W}\right),
\end{gather*}
where the network parameters contain two parts -- fixed pre-trained LLM weights $\mathbf{W}_0$ and trainable PEFT parameters $\Delta\mathbf{W}$. However, we have two distinct search spaces with architecture weights $\boldsymbol{\Theta}$ and $\boldsymbol{\Phi}$. These two kinds of architecture weights depend on each other, making optimizing them simultaneously hard. 

To address this issue, we propose an iterative optimization approach. 
We first use the first-order approximation in DARTS to optimize $\Delta\mathbf{W}$, which improves the search efficiency by considering that $\Delta\mathbf{W}$ converges fast based on the pre-trained $\mathbf{W}_0$.
After obtaining $\Delta\mathbf{W}^*$, we then fix the dimension parameters $\boldsymbol{\Phi}^*$ and optimize the module parameters $\boldsymbol{\Theta}$ first. Subsequently, we use the optimized $\boldsymbol{\Theta}^*$ to learn the optimal dimension parameters $\boldsymbol{\Phi}^*$. We repeat the previous steps until \ours converges. 

Besides, to bridge the gap between the search and validation stages, we follow~\cite{chang2019data} by employing the Gumbel-Softmax~\cite{jang2017categorical} function to normalize $\boldsymbol{\Theta}^*$ and $\boldsymbol{\Phi}^*$, where the nodes with maximal weight will have the highest probability to be selected. In addition, during the search, we implement weight entanglement, as utilized in AutoFormer~\cite{chen2021autoformer}, allowing shared weights across different dimensions to enhance stability, promoting faster convergence and reducing memory costs.

\begin{table*}[t]
\centering
\caption{Results of ablation study within the \textbf{S1}. ``\textbf{Entanglement}'' means that we directly use differential NAS to search from the entangled search space. ``$\textbf{b}\to \textbf{d}$'' is a two-stage search, where we first run a binary search until the model converges, then run the dimension rank search based on the binary search results. ``$\textbf{d}\to \textbf{b}$'' means to exchange the search order of ``$\textbf{b}\to \textbf{d}$''. ``\textbf{\textit{w.o.} Selection}'' is a variant of \ours by removing the early selection and only conducting the iterative search. ``\textbf{\textit{w.} Selection}'' is \ours.\colorbox{red!20}{Red} and \colorbox{green!20}{Green} highlight the best and second-best.} 
\vspace{-0.1in}
\resizebox{1\textwidth}{!}
{
\begin{tabular}{c|c|c|c|c|c|c|c|c|c}
\toprule
\multirow{3}{*}{\textbf{Entanglement}} & \multicolumn{4}{c|}{\textbf{Disentanglement}} & \multirow{3}{*}{\textbf{Ratio}} & \multirow{3}{*}{\textbf{COLA}} & \multirow{3}{*}{\textbf{MRPC}} & \multirow{3}{*}{\textbf{RTE}} & \multirow{3}{*}{\textbf{AVG}}\\\cline{2-5}

& \multicolumn{2}{c|}{\textbf{\textit{w.o.} Iteration}} & \multicolumn{2}{c|}{\textbf{\textit{w.} Iteration}} & & & & & \\\cline{2-5}
& $\textbf{b}\to \textbf{d}$ & $\textbf{d}\to \textbf{b}$ & \textbf{\textit{w.o.} Selection} & \textbf{\textit{w.} Selection} & & & & &  \\\hline
\checkmark & & & & & 33.81\%\% & 60.52 ± 2.58 & 91.54 ± 2.48 & 87.77 ± 1.01 & 79.94  \\\hline
 & \checkmark& & & & 29.01\%\%  &60.03 ± 2.94 & 91.66 ± 1.89  &88.01 ± 0.83 & 79.90  \\\hline
& & \checkmark& & & \cellcolor{green!20}21.98\%\% &60.33 ± 3.70 & 92.03 ± 1.36 & 87.77 ± 1.86 & 80.04  \\\hline
& & & \checkmark& & 29.22\%\% & \cellcolor{green!20}62.95 ± 3.03 & \cellcolor{green!20}92.04 ± 2.29 & \cellcolor{green!20}88.25 ± 0.83 &\cellcolor{green!20}81.08 \\\hline
& & & &\checkmark &\cellcolor{red!20}\textbf{1.39\%\%} & \cellcolor{red!20}\textbf{62.97 ± 3.72} & \cellcolor{red!20}\textbf{93.22 ± 0.62} & \cellcolor{red!20}\textbf{88.49 ± 0.83} &\cellcolor{red!20}\textbf{81.56} \\

\bottomrule
\end{tabular}}
\vspace{-0.2in}

\label{tab:ablation}

\end{table*}

\section{Experiments}








\subsection{Experimental Setups}

\noindent\ul{\textbf{Datasets.}}
We use two widely used natural language processing (NLP) benchmarks: GLUE ~\cite{wang-etal-2018-glue} and SuperGLUE ~\cite{wang2019superglue} in our experiments. All datasets are downloaded from the HuggingFace Datasets~\cite{lhoest-etal-2021-datasets}. 

We follow S$^3$Delta~\cite{hu2022sparse} to generate the training, validation, and test splits. For larger datasets, including QQP, QNLI, ReCoRD, SST-2, and MNLI, we allocate 2,000 random samples from the training set to form a new validation set, use the remaining samples as the training set, and repurpose the original validation set as the test set. For smaller datasets, we equally split the original validation set into new validation and test sets, while the original training set remains unchanged. Each dataset is split using different random seeds to introduce variability in the dataset configurations. Note that we remove the COPA dataset in SuperGLUE since its performance varies dramatically following S$^3$Delta~\cite{hu2022sparse}.

\smallskip
\noindent\ul{\textbf{Baselines.}}
We follow existing studies and use the following models as baselines:
(1) Automatic PEFT methods: 
\textbf{AutoPEFT}~\cite{zhou-etal-2024-autopeft} uses multi-objective Bayesian optimization to discover a Pareto-optimal set of PEFT configurations. 
\textbf{S$^3$Delta}~\cite{hu2022sparse} automatically searches sparse PEFT structure by determining the usage of each module using differential NAS.  
\textbf{PrunePEFT}~\cite{lawton-etal-2023-neural} implements a simple unstructured pruning strategy to search for optimal PEFT structures.
(2) Manually designed PEFT methods: We use \textbf{LoRA}~\cite{hu2022lora}, \textbf{Adapter}~\cite{houlsby2019parameter}, \textbf{BitFit}, and \textbf{LNFit} as baselines following {S$^3$Delta}~\cite{hu2022sparse}. 

\smallskip
\noindent\ul{\textbf{Pretrained Backbone Model.}} 
Our experiments utilize the T5 large model for all the baselines and \ours, which contains approximately 770 million parameters. We freeze the pretrained parameters $\mathbf{W}_0$ across all PEFT settings. 

\smallskip
\noindent\ul{\textbf{Search Spaces.}} 
Since the search space of existing work is different, for a fair comparison, we follow existing work and make comparisons within the space that they use. 
\ul{Setting 1 (\textbf{S1})}: For AutoPEFT, we use a mixture of Serial Adapter~\cite{houlsby2019parameter}, Parallel Adapter~\cite{he2022towards}, and Prefix-Tuning~\cite{li-liang-2021-prefix}. \ul{Setting 2 (\textbf{S2})}: When comparing with S$^3$Delta and PrunePEFT, we use a mixture of LoRA, Adapter-LR, BitFit~\cite{zaken2021bitfit}, and LNFiT modules as the search space. 
For both S1 and S2, each module has a binary selection space $\{0,1\}$. For baseline S$^3$Delta, its dimension size is fixed, which is 1. For other approaches, the candidate dimensions for all the PEFT modules in both search space settings are $\textsf{Dim}=\{1,4,8\}$. 
Additional experiment settings and hyperparameters are listed in Appendix~\ref{append:hyperparam}.

\subsection{Main Results}
We report the results on GLUE and SuperGLUE in Table~\ref{tab:results} by averaging 3 runs with different random seeds. We follow previous work~\cite{hu2022sparse} and use different metrics to validate the performance of different tasks, as detailed in Appendix~\ref{append:metrics}. Notably, \ours typically outperforms both manual PEFT methods and automated PEFT baselines regarding average scores on both benchmarks under different search spaces, with a small budget. On the GLUE benchmark, \ours even surpasses the full fine-tuning with only 1.39\%\% parameters.

When comparing the automated PEFT methods with manually designed PEFT, we find that \ours and other automated PEFT methods achieve better average scores on GLUE and SuperGLUE with fewer parameters. We also find that the differences in search space design result in performance change. Under the \textbf{S1} setting designed by AutoPEFT, \ours searched with an equivalent low parameter budget of 1.39\%\% does not match the performance achieved under the \textbf{S2} setting, which contains more types of PEFT modules, highlighting the benefits of including a diverse range of searchable PEFT modules in \ours.

\subsection{Ablation Study}

The proposed \ours mainly considers the disentanglement search spaces and uses both binary module and dimension rank early selection strategies to enhance efficiency and boost performance. We use four baselines in the ablation study to validate the effectiveness of our model design. The experimental setting follows~\cite{hu2022sparse}, and the results of these baselines are shown in Table~\ref{tab:ablation}. We can observe that the iterative search approaches (the last two rows) outperform the entanglement and non-iterative ones, indicating the effectiveness of our proposed iterative search solution. Besides, using the designed selection strategies, \ours further enhances its performance and reduces the number of parameters. This comparison validates the usefulness of our early selection strategies.

\vspace{-1mm}
\subsection{Efficiency Analysis}
In addition to the promising performance, we also evaluate the search efficiency of our method by comparing the time consumed during the search stage against the re-training duration, as detailed in Table~\ref{tab:efficiency}. All the training time is tested with the same batch size on the same GPU devices. The time is averaged from three datasets, RTE, STSB, and CoLA, under two search space settings, S1 and S2. It clearly shows that \ours achieves a high search efficiency with the design of early selection.

\begin{table}[t]
\centering
\caption{Efficiency Comparison.}
\label{tab:efficiency}
\vspace{-0.1in}
\resizebox{0.85\columnwidth}{!}{
\begin{tabular}{c|c|ccc}
\toprule
\multirow{2}{*}{\textbf{Setting}} & \multirow{2}{*}{\textbf{Method}} & \multicolumn{3}{c}{\textbf{Avg. Time (min)}} \\\cline{3-5}
                 &                 & \textbf{Search} & \textbf{Re-train} & \textbf{Ratio} \\
\midrule
\multirow{2}{*}{\textbf{S1}} & AutoPEFT   & 348 & 22 & 15.80 \\
                    & \ours & 13  & 22 & \cellcolor{red!20}\textbf{0.59} \\
\midrule
\multirow{3}{*}{\textbf{S2}} & $S^3$Delta & 125 & 20 & 6.30 \\
                    & PrunePEFT  & 20  & 20 & 1.00 \\
                    & \ours & 13  & 20 & \cellcolor{green!20}\textbf{0.65} \\
\bottomrule
\end{tabular}}
\end{table}

\begin{figure}[t]
  \centering
  \includegraphics[width=0.8\linewidth]{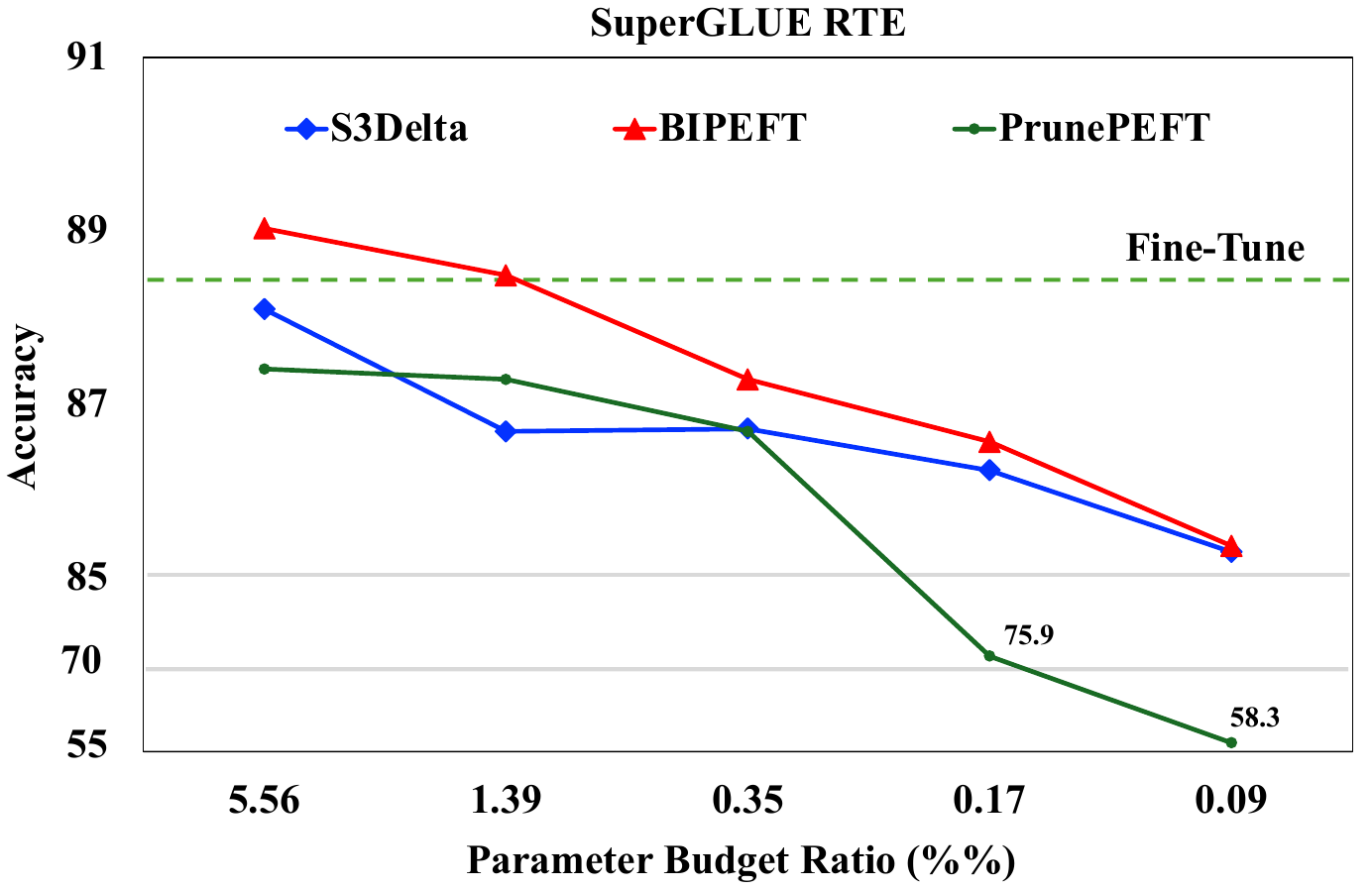}
  \caption{Performance vs. different levels of budget.}
  \vspace{-0.1 in}
  \label{fig:budget}
\end{figure}

\subsection{Different Parameter Budgets}
We also explore the influence of the parameter budget. Ideally, a larger budget will lead to better performance. Figure~\ref{fig:budget} shows the results with different ratios of parameter budgets, and the models are searched in space \textbf{S1} on the RTE dataset in the SuperGLUE benchmark, compared with S$^3$Delta and PrunePEFT. We can observe that \ours consistently sustains a higher accuracy and saturates to full fine-tuning performance by preserving essential PEFT modules even at very low budget levels.

\subsection{Generalization Ability Analysis}
We evaluate the task generalization capability of the structures searched within search space \textbf{S1}, as presented in Table~\ref{tab:transfer}. The results indicate that the structures searched for source datasets such as MRPC, MNLI, and QQP exhibit robust generalization to various target datasets, which encompass different types of NLP tasks. This demonstrates the ability of our searched structures in maintaining high performance across diverse NLP downstream applications.

\begin{table}[t]

\centering
\caption{PEFT structure generalization from source datasets to target
datasets. ``No Transfer'' means using the original structures searched from the target datasets.}
\label{tab:transfer}
\vspace{-0.1in}
\resizebox{1\columnwidth}{!}{
\begin{tabular}{c|cccc}
\toprule
 \textbf{Source} & \multicolumn{4}{c}{\textbf{Target Datasets}} \\
 \cmidrule(lr){2-5}
 \textbf{Datasets} & \textbf{STSB} & \textbf{SST2} & \textbf{QNLI} & \textbf{COLA} \\
\midrule
 \textbf{No Transfer} & \cellcolor{green!20}92.23 ± 0.28 & \cellcolor{red!20}96.27 ± 0.24 & \cellcolor{red!20}94.34 ± 0.06 & \cellcolor{red!20}62.95 ± 3.03  \\
 \textbf{MRPC} & \cellcolor{red!20}92.28 ± 0.22 & 96.04 ± 0.08 & \cellcolor{green!20}94.18 ± 0.16 &  60.51 ± 6.21 \\
 \textbf{MNLI} & 92.04 ± 0.12 & \cellcolor{green!20}96.16 ± 0.24 & 94.04 ± 0.17 & 60.17 ± 2.96  \\
 \textbf{QQP} & 92.20 ± 0.05 & 96.10 ± 0.16 & 94.16 ± 0.12 &  \cellcolor{green!20}62.08 ± 5.72\\
\bottomrule
\end{tabular}}
\vspace{-0.2in}
\end{table}



\section{Conclusion}
In this paper, we introduce \ours, a highly efficient search framework for parameter-efficient fine-tuning (PEFT) modules on large pretrained language models. \ours operates within a specifically designed disentangled search space using an iterative search strategy, incorporating novel selection mechanisms that significantly accelerate the search process while delivering promising performance outcomes.
Our extensive experiments on two widely used NLP benchmarks demonstrate the superiority of the PEFT structures identified by \ours. Despite requiring limited parameters and incurring minimal search costs, \ours outperforms both manually designed and other automated PEFT methods across a variety of NLP tasks. Additionally, the searched structures exhibit excellent generalization ability, proving effective for other downstream tasks as well.
Overall, \ours represents a substantial advancement in the field of automatic PEFT optimization, providing an efficient and effective solution for fine-tuning large pretrained language models. 


\section*{Limitations}
While we conduct our PEFT within the S1 and S2 search space settings, which include many popular PEFT modules, there remains the potential to integrate additional existing or new PEFT modules into our framework to further evaluate search performance and efficiency. Nonetheless, the proposed \ours framework is inherently flexible, allowing for the seamless integration of new PEFT modules as they become available. In addition, although the early stopping design requires manual hyperparameters like the maximum step $Z$, this mechanism will sustain high efficiency and strong performance because it could effectively prioritize the most critical PEFT modules within the given budget.

\section*{Acknowledgements}
This work is partially supported by the National Science Foundation under Grant No. 2238275 and 2212323 and the National Institutes of Health under Grant No. R01AG077016.



\bibliography{custom}

\begin{thebibliography}{42}
\expandafter\ifx\csname natexlab\endcsname\relax\def\natexlab#1{#1}\fi

\bibitem[{Ben~Zaken et~al.(2022)Ben~Zaken, Goldberg, and Ravfogel}]{zaken2021bitfit}
Elad Ben~Zaken, Yoav Goldberg, and Shauli Ravfogel. 2022.
\newblock {B}it{F}it: Simple parameter-efficient fine-tuning for transformer-based masked language-models.
\newblock In \emph{Proceedings of the 60th Annual Meeting of the Association for Computational Linguistics}, pages 1--9.

\bibitem[{Chang et~al.(2019)Chang, Guo, Meng, Xiang, Pan et~al.}]{chang2019data}
Jianlong Chang, Yiwen Guo, Gaofeng Meng, Shiming Xiang, Chunhong Pan, et~al. 2019.
\newblock Data: Differentiable architecture approximation.
\newblock \emph{NeurIPS}, 32.

\bibitem[{Chen et~al.(2023)Chen, Zhang, Shi, Li, Smola, and Yang}]{chen2023parameterefficient}
Jiaao Chen, Aston Zhang, Xingjian Shi, Mu~Li, Alex Smola, and Diyi Yang. 2023.
\newblock \href {https://openreview.net/forum?id=XSRSWxyJIC} {Parameter-efficient fine-tuning design spaces}.
\newblock In \emph{The Eleventh International Conference on Learning Representations}.

\bibitem[{Chen et~al.(2021)Chen, Peng, Fu, and Ling}]{chen2021autoformer}
Minghao Chen, Houwen Peng, Jianlong Fu, and Haibin Ling. 2021.
\newblock Autoformer: Searching transformers for visual recognition.
\newblock In \emph{ICCV}, pages 12270--12280.

\bibitem[{Dettmers et~al.(2023)Dettmers, Pagnoni, Holtzman, and Zettlemoyer}]{dettmers2023qlora}
Tim Dettmers, Artidoro Pagnoni, Ari Holtzman, and Luke Zettlemoyer. 2023.
\newblock Qlora: Efficient finetuning of quantized llms.
\newblock \emph{NeurIPS}, 36.

\bibitem[{Devlin et~al.(2019)Devlin, Chang, Lee, and Toutanova}]{Devlin2019BERTPO}
Jacob Devlin, Ming-Wei Chang, Kenton Lee, and Kristina Toutanova. 2019.
\newblock \href {https://api.semanticscholar.org/CorpusID:52967399} {Bert: Pre-training of deep bidirectional transformers for language understanding}.
\newblock In \emph{North American Chapter of the Association for Computational Linguistics}.

\bibitem[{Dosovitskiy et~al.(2021)Dosovitskiy, Beyer, Kolesnikov, Weissenborn, Zhai, Unterthiner, Dehghani, Minderer, Heigold, Gelly, Uszkoreit, and Houlsby}]{dosovitskiy2021an}
Alexey Dosovitskiy, Lucas Beyer, Alexander Kolesnikov, Dirk Weissenborn, Xiaohua Zhai, Thomas Unterthiner, Mostafa Dehghani, Matthias Minderer, Georg Heigold, Sylvain Gelly, Jakob Uszkoreit, and Neil Houlsby. 2021.
\newblock \href {https://openreview.net/forum?id=YicbFdNTTy} {An image is worth 16x16 words: Transformers for image recognition at scale}.
\newblock In \emph{International Conference on Learning Representations}.

\bibitem[{Fu et~al.(2023)Fu, Yang, So, Lam, Bing, and Collier}]{fu2023effectiveness}
Zihao Fu, Haoran Yang, Anthony Man-Cho So, Wai Lam, Lidong Bing, and Nigel Collier. 2023.
\newblock On the effectiveness of parameter-efficient fine-tuning.
\newblock In \emph{Proceedings of the AAAI Conference on Artificial Intelligence}, volume~37, pages 12799--12807.

\bibitem[{Guo et~al.(2021)Guo, Rush, and Kim}]{guo-etal-2021-parameter}
Demi Guo, Alexander Rush, and Yoon Kim. 2021.
\newblock Parameter-efficient transfer learning with diff pruning.
\newblock In \emph{Proceedings of the 59th Annual Meeting of the Association for Computational Linguistics and the 11th International Joint Conference on Natural Language Processing (Volume 1: Long Papers)}, pages 4884--4896.

\bibitem[{Han et~al.(2024)Han, Gao, Liu, Zhang et~al.}]{han2024parameter}
Zeyu Han, Chao Gao, Jinyang Liu, Sai~Qian Zhang, et~al. 2024.
\newblock Parameter-efficient fine-tuning for large models: A comprehensive survey.
\newblock \emph{arXiv preprint arXiv:2403.14608}.

\bibitem[{He et~al.(2022)He, Zhou, Ma, Berg-Kirkpatrick, and Neubig}]{he2022towards}
Junxian He, Chunting Zhou, Xuezhe Ma, Taylor Berg-Kirkpatrick, and Graham Neubig. 2022.
\newblock \href {https://openreview.net/forum?id=0RDcd5Axok} {Towards a unified view of parameter-efficient transfer learning}.
\newblock In \emph{International Conference on Learning Representations}.

\bibitem[{Houlsby et~al.(2019)Houlsby, Giurgiu, Jastrzebski, Morrone, De~Laroussilhe, Gesmundo, Attariyan, and Gelly}]{houlsby2019parameter}
Neil Houlsby, Andrei Giurgiu, Stanislaw Jastrzebski, Bruna Morrone, Quentin De~Laroussilhe, Andrea Gesmundo, Mona Attariyan, and Sylvain Gelly. 2019.
\newblock Parameter-efficient transfer learning for nlp.
\newblock In \emph{International Conference on Machine Learning}, pages 2790--2799. PMLR.

\bibitem[{Hu et~al.(2022{\natexlab{a}})Hu, yelong shen, Wallis, Allen-Zhu, Li, Wang, Wang, and Chen}]{hu2022lora}
Edward~J Hu, yelong shen, Phillip Wallis, Zeyuan Allen-Zhu, Yuanzhi Li, Shean Wang, Lu~Wang, and Weizhu Chen. 2022{\natexlab{a}}.
\newblock \href {https://openreview.net/forum?id=nZeVKeeFYf9} {Lo{RA}: Low-rank adaptation of large language models}.
\newblock In \emph{International Conference on Learning Representations}.

\bibitem[{Hu et~al.(2022{\natexlab{b}})Hu, Zhang, Ding, Wang, Wang, Liu, and Sun}]{hu2022sparse}
Shengding Hu, Zhen Zhang, Ning Ding, Yadao Wang, Yasheng Wang, Zhiyuan Liu, and Maosong Sun. 2022{\natexlab{b}}.
\newblock \href {https://openreview.net/forum?id=oOte_397Q4P} {Sparse structure search for delta tuning}.
\newblock In \emph{Thirty-Sixth Conference on Neural Information Processing Systems}.

\bibitem[{Jang et~al.(2017)Jang, Gu, and Poole}]{jang2017categorical}
Eric Jang, Shixiang Gu, and Ben Poole. 2017.
\newblock \href {https://openreview.net/forum?id=rkE3y85ee} {Categorical reparameterization with gumbel-softmax}.
\newblock In \emph{International Conference on Learning Representations}.

\bibitem[{Jiang et~al.(2023)Jiang, Ji, Zhu, Yuan, and Huang}]{jiang2023operationlevel}
Shen Jiang, Zipeng Ji, Guanghui Zhu, Chunfeng Yuan, and Yihua Huang. 2023.
\newblock \href {https://openreview.net/forum?id=yAOwkf4FyL} {Operation-level early stopping for robustifying differentiable {NAS}}.
\newblock In \emph{Thirty-seventh Conference on Neural Information Processing Systems}.

\bibitem[{Lawton et~al.(2023)Lawton, Kumar, Thattai, Galstyan, and Ver~Steeg}]{lawton-etal-2023-neural}
Neal Lawton, Anoop Kumar, Govind Thattai, Aram Galstyan, and Greg Ver~Steeg. 2023.
\newblock Neural architecture search for parameter-efficient fine-tuning of large pre-trained language models.
\newblock In \emph{Findings of the Association for Computational Linguistics: ACL 2023}, pages 8506--8515.

\bibitem[{Lhoest et~al.(2021)Lhoest, Villanova~del Moral, Jernite, Thakur, von Platen, Patil, Chaumond, Drame, Plu, Tunstall, Davison, {\v{S}}a{\v{s}}ko, Chhablani, Malik, Brandeis, Le~Scao, Sanh, Xu, Patry, McMillan-Major, Schmid, Gugger, Delangue, Matussi{\`e}re, Debut, Bekman, Cistac, Goehringer, Mustar, Lagunas, Rush, and Wolf}]{lhoest-etal-2021-datasets}
Quentin Lhoest, Albert Villanova~del Moral, Yacine Jernite, Abhishek Thakur, Patrick von Platen, Suraj Patil, Julien Chaumond, Mariama Drame, Julien Plu, Lewis Tunstall, Joe Davison, Mario {\v{S}}a{\v{s}}ko, Gunjan Chhablani, Bhavitvya Malik, Simon Brandeis, Teven Le~Scao, Victor Sanh, Canwen Xu, Nicolas Patry, Angelina McMillan-Major, Philipp Schmid, Sylvain Gugger, Cl{\'e}ment Delangue, Th{\'e}o Matussi{\`e}re, Lysandre Debut, Stas Bekman, Pierric Cistac, Thibault Goehringer, Victor Mustar, Fran{\c{c}}ois Lagunas, Alexander Rush, and Thomas Wolf. 2021.
\newblock Datasets: A community library for natural language processing.
\newblock In \emph{Proceedings of the 2021 Conference on Empirical Methods in Natural Language Processing: System Demonstrations}, pages 175--184.

\bibitem[{Li and Liang(2021)}]{li-liang-2021-prefix}
Xiang~Lisa Li and Percy Liang. 2021.
\newblock Prefix-tuning: Optimizing continuous prompts for generation.
\newblock In \emph{Proceedings of the 59th Annual Meeting of the Association for Computational Linguistics and the 11th International Joint Conference on Natural Language Processing (Volume 1: Long Papers)}, pages 4582--4597.

\bibitem[{Liang et~al.(2019)Liang, Zhang, Sun, He, Huang, Zhuang, and Li}]{liang2019darts+}
Hanwen Liang, Shifeng Zhang, Jiacheng Sun, Xingqiu He, Weiran Huang, Kechen Zhuang, and Zhenguo Li. 2019.
\newblock Darts+: Improved differentiable architecture search with early stopping.
\newblock \emph{arXiv preprint arXiv:1909.06035}.

\bibitem[{Liao et~al.(2023)Liao, Meng, and Monz}]{liao-etal-2023-parameter}
Baohao Liao, Yan Meng, and Christof Monz. 2023.
\newblock Parameter-efficient fine-tuning without introducing new latency.
\newblock In \emph{Proceedings of the 61st Annual Meeting of the Association for Computational Linguistics}, pages 4242--4260.

\bibitem[{Lin et~al.(2020)Lin, Madotto, and Fung}]{lin2020exploring}
Zhaojiang Lin, Andrea Madotto, and Pascale Fung. 2020.
\newblock Exploring versatile generative language model via parameter-efficient transfer learning.
\newblock In \emph{Findings of the Association for Computational Linguistics: EMNLP 2020}, pages 441--459.

\bibitem[{Liu et~al.(2019)Liu, Simonyan, and Yang}]{liu2018darts}
Hanxiao Liu, Karen Simonyan, and Yiming Yang. 2019.
\newblock \href {https://openreview.net/forum?id=S1eYHoC5FX} {{DARTS}: Differentiable architecture search}.
\newblock In \emph{International Conference on Learning Representations}.

\bibitem[{Luo et~al.(2023)Luo, Huang, Zhou, Sun, Jiang, Wang, and Ji}]{luo2023towards}
Gen Luo, Minglang Huang, Yiyi Zhou, Xiaoshuai Sun, Guannan Jiang, Zhiyu Wang, and Rongrong Ji. 2023.
\newblock Towards efficient visual adaption via structural re-parameterization.
\newblock \emph{arXiv preprint arXiv:2302.08106}.

\bibitem[{Luo et~al.(2024)Luo, Wang, Wang, Chang, Wang, and Ma}]{luo-etal-2024-corelation}
Junyu Luo, Xiaochen Wang, Jiaqi Wang, Aofei Chang, Yaqing Wang, and Fenglong Ma. 2024.
\newblock \href {https://aclanthology.org/2024.lrec-main.355} {{C}o{R}elation: Boosting automatic {ICD} coding through contextualized code relation learning}.
\newblock In \emph{LREC-COLING 2024}, pages 3997--4007, Torino, Italia. ELRA and ICCL.

\bibitem[{Mao et~al.(2021)Mao, Mathias, Hou, Almahairi, Ma, Han, tau Yih, and Khabsa}]{Mao2021UniPELTAU}
Yuning Mao, Lambert Mathias, Rui Hou, Amjad Almahairi, Hao Ma, Jiawei Han, Wen tau Yih, and Madian Khabsa. 2021.
\newblock \href {https://api.semanticscholar.org/CorpusID:238857301} {Unipelt: A unified framework for parameter-efficient language model tuning}.
\newblock In \emph{Annual Meeting of the Association for Computational Linguistics}.

\bibitem[{Molchanov et~al.(2019)Molchanov, Mallya, Tyree, Frosio, and Kautz}]{molchanov2019importance}
Pavlo Molchanov, Arun Mallya, Stephen Tyree, Iuri Frosio, and Jan Kautz. 2019.
\newblock Importance estimation for neural network pruning.
\newblock In \emph{Proceedings of the IEEE/CVF conference on computer vision and pattern recognition}, pages 11264--11272.

\bibitem[{Radford et~al.(2019)Radford, Wu, Child, Luan, Amodei, Sutskever et~al.}]{radford2019language}
Alec Radford, Jeffrey Wu, Rewon Child, David Luan, Dario Amodei, Ilya Sutskever, et~al. 2019.
\newblock Language models are unsupervised multitask learners.
\newblock \emph{OpenAI blog}, 1(8):9.

\bibitem[{R{\"u}ckl{\'e} et~al.(2021)R{\"u}ckl{\'e}, Geigle, Glockner, Beck, Pfeiffer, Reimers, and Gurevych}]{ruckle2021adapterdrop}
Andreas R{\"u}ckl{\'e}, Gregor Geigle, Max Glockner, Tilman Beck, Jonas Pfeiffer, Nils Reimers, and Iryna Gurevych. 2021.
\newblock Adapterdrop: On the efficiency of adapters in transformers.
\newblock In \emph{Proceedings of the 2021 Conference on Empirical Methods in Natural Language Processing}, pages 7930--7946.

\bibitem[{Su et~al.(2023)Su, Chan, Cheng, Qin, Lin, Hu, Yang, Ding, Sun, Xie et~al.}]{su2023exploring}
Yusheng Su, Chi-Min Chan, Jiali Cheng, Yujia Qin, Yankai Lin, Shengding Hu, Zonghan Yang, Ning Ding, Xingzhi Sun, Guotong Xie, et~al. 2023.
\newblock Exploring the impact of model scaling on parameter-efficient tuning.
\newblock In \emph{Proceedings of the 2023 Conference on Empirical Methods in Natural Language Processing}, pages 15062--15078.

\bibitem[{Sung et~al.(2021)Sung, Nair, and Raffel}]{sung2021training}
Yi-Lin Sung, Varun Nair, and Colin~A Raffel. 2021.
\newblock Training neural networks with fixed sparse masks.
\newblock \emph{NeurIPS}, 34:24193--24205.

\bibitem[{Vaswani et~al.(2017)Vaswani, Shazeer, Parmar, Uszkoreit, Jones, Gomez, Kaiser, and Polosukhin}]{vaswani2017attention}
Ashish Vaswani, Noam Shazeer, Niki Parmar, Jakob Uszkoreit, Llion Jones, Aidan~N Gomez, {\L}ukasz Kaiser, and Illia Polosukhin. 2017.
\newblock Attention is all you need.
\newblock \emph{NeurIPS}, 30.

\bibitem[{Wang et~al.(2019)Wang, Pruksachatkun, Nangia, Singh, Michael, Hill, Levy, and Bowman}]{wang2019superglue}
Alex Wang, Yada Pruksachatkun, Nikita Nangia, Amanpreet Singh, Julian Michael, Felix Hill, Omer Levy, and Samuel Bowman. 2019.
\newblock Superglue: A stickier benchmark for general-purpose language understanding systems.
\newblock \emph{NeurIPS}, 32.

\bibitem[{Wang et~al.(2018)Wang, Singh, Michael, Hill, Levy, and Bowman}]{wang-etal-2018-glue}
Alex Wang, Amanpreet Singh, Julian Michael, Felix Hill, Omer Levy, and Samuel Bowman. 2018.
\newblock {GLUE}: A multi-task benchmark and analysis platform for natural language understanding.
\newblock In \emph{Proceedings of the 2018 {EMNLP} Workshop {B}lackbox{NLP}: Analyzing and Interpreting Neural Networks for {NLP}}, pages 353--355.

\bibitem[{Wang et~al.(2024)Wang, Luo, Ye, Wang, Zhong, Chang, Huang, Yin, Xiao, Sun, and Ma}]{ijcai2024p914}
Jiaqi Wang, Junyu Luo, Muchao Ye, Xiaochen Wang, Yuan Zhong, Aofei Chang, Guanjie Huang, Ziyi Yin, Cao Xiao, Jimeng Sun, and Fenglong Ma. 2024.
\newblock \href {https://doi.org/10.24963/ijcai.2024/914} {Recent advances in predictive modeling with electronic health records}.
\newblock In \emph{Proceedings of the Thirty-Third International Joint Conference on Artificial Intelligence, {IJCAI-24}}, pages 8272--8280. International Joint Conferences on Artificial Intelligence Organization.
\newblock Survey Track.

\bibitem[{Xu et~al.(2023)Xu, Xie, Qin, Tao, and Wang}]{xu2023parameter}
Lingling Xu, Haoran Xie, Si-Zhao~Joe Qin, Xiaohui Tao, and Fu~Lee Wang. 2023.
\newblock Parameter-efficient fine-tuning methods for pretrained language models: A critical review and assessment.
\newblock \emph{arXiv preprint arXiv:2312.12148}.

\bibitem[{Zhang et~al.(2023{\natexlab{a}})Zhang, Li, Chen, Jiang, Wang, and Qian}]{zhang2023increlora}
Feiyu Zhang, Liangzhi Li, Junhao Chen, Zhouqiang Jiang, Bowen Wang, and Yiming Qian. 2023{\natexlab{a}}.
\newblock Increlora: Incremental parameter allocation method for parameter-efficient fine-tuning.
\newblock \emph{arXiv preprint arXiv:2308.12043}.

\bibitem[{Zhang et~al.(2023{\natexlab{b}})Zhang, Chen, Bukharin, He, Cheng, Chen, and Zhao}]{zhang2023adaptive}
Qingru Zhang, Minshuo Chen, Alexander Bukharin, Pengcheng He, Yu~Cheng, Weizhu Chen, and Tuo Zhao. 2023{\natexlab{b}}.
\newblock Adaptive budget allocation for parameter-efficient fine-tuning.
\newblock In \emph{International Conference on Learning Representations}. Openreview.

\bibitem[{Zhang et~al.(2022{\natexlab{a}})Zhang, Zuo, Liang, Bukharin, He, Chen, and Zhao}]{zhang2022platon}
Qingru Zhang, Simiao Zuo, Chen Liang, Alexander Bukharin, Pengcheng He, Weizhu Chen, and Tuo Zhao. 2022{\natexlab{a}}.
\newblock Platon: Pruning large transformer models with upper confidence bound of weight importance.
\newblock In \emph{International conference on machine learning}, pages 26809--26823. PMLR.

\bibitem[{Zhang et~al.(2022{\natexlab{b}})Zhang, Zhou, and Liu}]{zhang2022neural}
Yuanhan Zhang, Kaiyang Zhou, and Ziwei Liu. 2022{\natexlab{b}}.
\newblock \href {http://arxiv.org/abs/2206.04673} {Neural prompt search}.

\bibitem[{Zhou et~al.(2024)Zhou, Wan, Vuli{\'c}, and Korhonen}]{zhou-etal-2024-autopeft}
Han Zhou, Xingchen Wan, Ivan Vuli{\'c}, and Anna Korhonen. 2024.
\newblock Autopeft: Automatic configuration search for parameter-efficient fine-tuning.
\newblock \emph{Transactions of the Association for Computational Linguistics}, 12.

\bibitem[{Zi et~al.(2023)Zi, Qi, Wang, Wang, Wong, and Zhang}]{zi2023delta}
Bojia Zi, Xianbiao Qi, Lingzhi Wang, Jianan Wang, Kam-Fai Wong, and Lei Zhang. 2023.
\newblock Delta-lora: Fine-tuning high-rank parameters with the delta of low-rank matrices.
\newblock \emph{arXiv preprint arXiv:2309.02411}.

\end{thebibliography}
\bibliographystyle{acl_natbib}

\appendix

\begin{table*}[]
\centering
\caption{Additional Results on the GLUE benchmark.$*$ denotes the results that are directly copied from \textbf{S$^3$Delta}~\cite{hu2022sparse}. 
\colorbox{red!20}{Red} denotes the best score.
}
\label{tab:appendix_higher_ratio}
\vspace{-0.1in}
\resizebox{1\textwidth}{!}{
\begin{tabular}{l|l|ccccccc|c} 
\toprule 
\multicolumn{10}{c}{\cellcolor{gray!10}\textbf{GLUE}} \\\hline
     \textbf{Ratio}
        & \textbf{Method} & \textbf{CoLA} & \textbf{SST2} & \textbf{MRPC} & \textbf{QQP} & \textbf{STSB} & \textbf{MNLI} &  \textbf{QNLI} & \textbf{AVG} \\  
        \midrule
        
    10000\%\% & Fine-tune$^*$  & 62.25 ± 3.96 & 95.87 ± 0.42 & 91.86 ± 1.19 & 89.50 ± 0.22 & 91.86 ± 0.46 & 89.61 ± 0.30 & 94.22 ± 0.35 & 87.88 \\ 
    \midrule
    \multicolumn{10}{c}{\textbf{Automated PEFT Methods}} \\\hline
     5.56\%\% & $\text{S}^3$Delta-M$^*$& 61.67 ± 5.38 & 95.88 ± 0.11 & 91.16 ± 2.09 & 88.22 ± 0.04 & 91.81 ± 0.41 & 89.18 ± 0.08 & 93.88 ± 0.05 & 87.40 \\
     5.56\%\% & PrunePEFT & 63.83 ± 2.32 & 95.95 ± 0.19 & 91.67 ± 1.13 & \cellcolor{red!20}88.45 ± 0.08 & 92.06 ± 0.47 & \cellcolor{red!20}89.50 ± 0.14 & \cellcolor{red!20}94.35 ± 0.08 & 87.97 \\
     5.56\%\% & \ours & \cellcolor{red!20}64.48 ± 2.63 & \cellcolor{red!20}96.04 ± 0.40 & \cellcolor{red!20}93.19 ± 1.24 & 88.32 ± 0.14 & \cellcolor{red!20}92.09 ± 0.30 & 89.19 ± 0.02 & 94.25 ± 0.04 & \cellcolor{red!20}88.22 \\
    \bottomrule 
\end{tabular}
}
\vspace{-0.1in}

\end{table*}


\begin{table}[t]
\centering
\caption{The sensitivity analysis of the maximum trigger count $Z$. ``AVG.Time'' denotes the average search time (in minutes).}
\label{tab:hyper_z}
\vspace{-0.1in}
\resizebox{1\columnwidth}{!}{
\begin{tabular}{c|ccccc}
\toprule
 \textbf{Value of $Z$}& \textbf{COLA} & \textbf{MRPC} & \textbf{RTE} & \textbf{AVG} & \textbf{AVG. Time} \\
\midrule
 10 & 60.96 ± 1.22 & 92.98 ± 1.39 & 86.69 ± 0.36 & 80.21 & 6.34\\
 50 & 61.82 ± 3.85 & 92.78 ± 0.84 & 87.77 ± 1.44 & 80.79 & 8.41\\
 100 & 62.97 ± 3.72 & 93.22 ± 0.62 & 88.49 ± 2.04 & 81.56 & 12.39 \\
 200 & 62.99 ± 5.38 & 92.98 ± 1.09 & 89.21 ± 0.72 & 81.73 & 16.34\\
\bottomrule
\end{tabular}}
\vspace{-0.1in}
\end{table}

\begin{table}[]
\centering
\caption{The sensitivity analysis of $\gamma$.}
\label{tab:hyper_gamma}
\vspace{-0.1in}
\resizebox{1\columnwidth}{!}{
\begin{tabular}{c|cccc}
\toprule
 \textbf{Value of $\gamma$}& \textbf{COLA} & \textbf{MRPC} & \textbf{RTE} & \textbf{AVG} \\
\midrule
0.75 & 63.01 ± 2.86 & 93.24 ± 1.16 & 87.77 ± 0.72 & 81.34 \\
0.85 & 62.97 ± 3.72 & 93.22 ± 0.62 & 88.49 ± 2.04 & 81.56 \\
1 & 56.83 ± 4.70 & 92.01 ± 1.06 & 83.45 ± 0.59 & 77.43 \\
\bottomrule
\end{tabular}}
\vspace{-0.2in}
\end{table}

\begin{table}[]
\centering
\vspace{-0.15in}
\caption{The sensitivity analysis of time window $H$ used in the trigger design.}
\label{tab:hyper_H}
\vspace{-0.1in}
\resizebox{1\columnwidth}{!}{
\begin{tabular}{c|cccc}
\toprule
 \textbf{Value of $H$}& \textbf{COLA} & \textbf{MRPC} & \textbf{RTE} & \textbf{AVG} \\
\midrule
3 & 62.29 ± 4.80 & 93.26 ± 0.84 & 88.13 ± 0.36 & 81.23 \\
5 & 62.97 ± 3.72 & 93.22 ± 0.62 & 88.49 ± 2.04 & 81.56  \\
10 & 61.89 ± 4.54 & 93.45 ± 0.40 & 88.13 ± 0.36 & 81.16 \\
20 & 62.72 ± 3.99 & 92.64 ± 1.25 & 88.13 ± 1.08 & 81.16 \\
\bottomrule
\end{tabular}}
\end{table}

\begin{table}[]
\centering
\caption{The sensitivity analysis of stability threshold $\tau$.}
\label{tab:hyper_tau}
\vspace{-0.1in}
\resizebox{1\columnwidth}{!}{
\begin{tabular}{c|cccc}
\toprule
 \textbf{Value of $\tau$}& \textbf{COLA} & \textbf{MRPC} & \textbf{RTE} & \textbf{AVG} \\
\midrule
0.1 & 61.57 ± 4.62 & 92.74 ± 1.02 & 87.41 ± 0.36 & 80.57 \\
0.5 & 62.18 ± 4.10 & 92.94 ± 1.04 & 87.41 ± 0.36 & 80.84 \\
0.85 & 62.97 ± 3.72 & 93.22 ± 0.62 & 88.49 ± 2.04 & 81.56 \\
\bottomrule
\end{tabular}}
\vspace{-0.1in}
\end{table}

\begin{table}[]
\centering
\caption{Comparison with APET.}
\label{tab:compare_apet}
\vspace{-0.1in}
\resizebox{1\columnwidth}{!}{
\begin{tabular}{c|c|cccc}
\toprule
 \textbf{Method}& \textbf{Ratio} & \textbf{COLA} & \textbf{MRPC} & \textbf{RTE} & \textbf{AVG} \\
\midrule
 APET& 1.39\%\% & 59.99 ± 4.87 & 91.92 ± 0.67 & 85.61 ± 2.16 & 79.17\\
 APET& 5.56\%\% & 60.82 ± 3.95 & 92.68 ± 1.17 & 88.49 ± 1.01 & 80.66\\
 \ours& 1.39\%\% & 62.97 ± 3.72 & 93.22 ± 0.62 & 88.49 ± 2.04 & 81.56\\
\bottomrule
\end{tabular}}
\vspace{-0.2in}
\end{table}

\section{Evaluation Metrics}\label{append:metrics}
\label{appendix:exp_detail}

For both GLUE and SuperGLUE benchmarks, we employ various evaluation metrics: Accuracy is reported for the SST-2, MNLI, QNLI, BoolQ, CB, RTE, and WIC tasks. We utilize the F1 score to assess performance on MRPC, QQP, MultiRC, and ReCoRD. Additionally, Matthew's Correlation is used to evaluate CoLA, and the Pearson Correlation coefficient is applied to the STSB task.

\section{Experiment Settings and Hyperparameters}
\label{append:hyperparam}
For each experiment setting, we report the average performances and standard deviations using results from three different seeds on the final test sets. We configure the maximum sequence lengths as 128 for GLUE tasks and 256 for SuperGLUE tasks, maintaining a consistent batch size of 32 across both benchmarks. For the ReCoRD task, we adjust the settings to a maximum sequence length of 512 and a batch size of 16.
We utilize the AdamW optimizer with a linear learning rate decay schedule to optimize our model. All experiments maintain a consistent learning rate of \(3 \times 10^{-4}\) for PEFT training, while the learning rate for architecture parameters in \ours is set at 0.01.
For \ours, following DARTS~\cite{liu2018darts}, we equally split the original training set into two parts.
One part is used for optimizing the model parameters, and the other for optimizing the architecture parameters. The original validation set serves to evaluate and save the searched structures. For the default values of the model hyperparameters used in \ours, we set \(Z = 100\), \(\gamma = 0.85\), \(H = 5\), and \(\tau = 0.85\).

\section{Additional Experiment Results}

\subsection{Hyperparameter Sensitivity Analysis}

We perform a detailed hyperparameter sensitivity analysis using the same datasets as in our ablation study. Specifically, we test the following hyperparameters: (1) maximum trigger count $Z$, (2) $\gamma$, used for the smoothed sensitivity in Eq.~\eqref{eq:sensitivity}, (3) time window $H$, employed in the trigger design to measure model stability, and (4) stability threshold $\tau$. All these hyperparameters are configured for the sub-modules of our early selection module, which is designed to enhance search efficiency. The parameter budget ratios are uniformly set to 1.39\%\%. A comprehensive analysis of each hyperparameter is presented below. 

The maximum trigger count $Z$ controls the speed of the early selection process. We report the average search time across three datasets, demonstrating that while larger $Z$ values ensure smoother selection, they may marginally reduce efficiency. As shown in Table~\ref{tab:hyper_z}, increasing $Z$ beyond 100 diminishes marginal returns in performance improvement.
The parameter $\gamma$ is used for smoothed sensitivity in Eq.~\eqref{eq:sensitivity}, capturing the ratio of historical sensitivity information retained during the training process. The default value for $\gamma$ is set to 0.85. Our additional experiments, presented in Table~\ref{tab:hyper_gamma}, indicate that setting it to 1 prevents the sensitivity from being updated with new data. Stable performance is observed when $\gamma$ remains 0.7 to 0.9.
The time window $H$ is used in the trigger design to measure model stability. As indicated by the results in Table~\ref{tab:hyper_H}, varying $H$ does not produce significant performance differences across the three datasets. The default value of $H$ is set to 5.
The stability threshold $\tau$ influences trigger generation, with higher values enforcing stricter stability requirements, leading to smoother trigger generation and improved performance of the searched structures, as demonstrated in Table~\ref{tab:hyper_tau}. By default, we set $\tau$ to 0.85.

\subsection{Higher Sparsity Ratio of Parameters}
As demonstrated in the main experiments, \ours outperforms all baselines at a parameter ratio of 1.39\%\%. To further showcase the effectiveness of our method across diverse budget settings, we provide additional results on the GLUE benchmark at a parameter ratio of 5.56\%\% under setting \textbf{S2}, compared to PrunePEFT and $\text{S}^3$Delta. The average score for full fine-tuning is 87.88. As shown in Table~\ref{tab:appendix_higher_ratio}, \ours continues to outperform other baselines given a higher budget.

\subsection{Zero-Cost PEFT Configuration}
As introduced in~\cite{su2023exploring}, their zero-cost PEFT configuration method, APET, initially freezes key factors such as the number of trainable parameters. Under this constraint, APET arbitrarily selects tunable parameters using different random seeds, each representing a distinct parameter distribution, and trains them on the tasks. As shown in Table~\ref{tab:compare_apet}, \ours outperforms APET even at a lower parameter ratio, albeit with a small search cost.

\label{sec:appendix}

\end{document}